\documentclass[3p,times]{elsarticle}

\usepackage{ecrc}

\volume{00}

\firstpage{1}

\runauth{D. C. Wickramarachchi et al.}
\runauth{}

\usepackage{amssymb}

\usepackage{amsthm}
\usepackage[cmex10]{amsmath}
\usepackage{amsfonts}
\usepackage{amssymb}
\usepackage[boxed]{algorithm2e}
\usepackage{epstopdf} 
\usepackage{setspace}
\biboptions{longnamesfirst,sort}

\usepackage[figuresright]{rotating}

\begin{document}

\begin{frontmatter}



\author[UoC]{D. C. ˜Wickramarachchi\corref{*}}
\ead{wchitraka@gmail.com}

\author[UoW]{B. L. ˜Robertson}
\ead{brober25@uwyo.edu}

\author[UoC]{M. ~Reale}
\ead{Marco.Reale@canterbury.ac.nz}

\author[UoC]{C. J. ˜Price}
\ead{Chris.Price@canterbury.ac.nz}

\author[UoC]{J. ˜Brown}
\ead{Jennifer.Brown@canterbury.ac.nz}

\cortext[*]{Corresponding author}

\address[UoC]{School of Mathematics and Statistics, University of Canterbury, Private Bag 4800,
Christchurch, New Zealand 8042}

\address[UoW]{Department of Statistics, University of Wyoming, Laramie USA}


\title{HHCART: An Oblique Decision Tree}

\begin{abstract}
Decision trees are a popular technique in statistical data classification. They recursively partition the feature space into disjoint sub-regions until each sub-region becomes homogeneous with respect to a particular class. The basic Classification and Regression Tree (CART) algorithm partitions the feature space using axis parallel splits. When the true decision boundaries are not aligned with the feature axes, this approach can produce a complicated boundary structure. Oblique decision trees use oblique decision  boundaries to potentially simplify the boundary structure. The major limitation of this approach is that the tree induction algorithm is computationally expensive. In this article we present a new decision tree algorithm, called HHCART. The method utilizes a series of Householder matrices to reflect the training data at each node during the tree construction. Each reflection is based on the directions of the eigenvectors from each classes' covariance matrix. Considering axis parallel splits in the reflected training data provides an efficient way of finding oblique splits in the unreflected training data. Experimental results show that the accuracy and size of the HHCART trees are comparable with some benchmark methods in the literature. The appealing feature of HHCART is that it can handle both qualitative and quantitative features in the same oblique split.

\end{abstract}


\end{frontmatter}

\section{Introduction}

Decision trees (DTs) are an increasingly popular method used for classifying data. In the typical tree building procedure, the space that the data occupies (feature space) is iteratively partitioned into disjoint sub-regions until each sub-region is homogeneous (or near so) with respect to a particular class. In a DT, each sub-region is represented by a node in the tree. The node can be either terminal or non-terminal. Non-terminal nodes are impure and can be split further using a series of tests based on the feature variables, a process called splitting. Each split is determined by considering a series of hyperplanes which separate the feature space into two sub-regions. The best hyperplane split is chosen as the one which maximises the change in an impurity function ($\Delta(I)$). To obtain a fully grown tree, this process is recursively applied to each non-terminal node until terminal nodes are reached. The terminal nodes correspond to homogeneous or near homogeneous sub-regions in the feature space. Each terminal node is assigned the class label that minimises the misclassification cost at the node.\\
DTs play an important role in statistical learning and have been a popular technique for data classification over several decades (see \citep{breiman1984classification,murthy1994system,lopez2013fisher}). In the tree building process the aim is to produce accurate and smaller trees while minimising the computational time. Accuracy, size and time mainly depend on the way non-terminal nodes are split in a DT. Three types of splits are considered including axis parallel, oblique and non-linear splits. Axis parallel splits partition the space parallel to feature axes. Therefore axis parallel trees are desirable when the decision boundaries are aligned with the feature axes. Oblique splits are hyperplane splits defined by a linear combination of the feature variables. These splits are more appealing when the decision boundaries are not aligned with the feature axes. Non-linear splits \citep{ittner1996non, li2005classifiability} are general class of splits. Decision boundaries generated by these splits can take arbitrary shapes and can easily be influenced by noise data. \citep{li2005classifiability}.\\
Many algorithms have been proposed to induce DTs. In general, these algorithms differ in the way they search for the best split at each non-terminal node. Many studies show that trees which use oblique splits generally produce smaller trees with better accuracy compared with axis parallel trees \cite{li2003multivariate}. Therefore they have become increasingly popular in DT literature and motivated us to propose a new methodology to construct a DT which uses oblique splits at each non-terminal node. These DTs are called Oblique Decision Trees \citep{murthy1994system}. More specifically, let the feature vector consists of $p$ attributes, $\bf{x}$$=[\it{x_1},\it{x_2},\dots,\it{x_p}]^T$ where $\it{x_i} \in \mathbb{R}$. The oblique splits can be defined as linear combinations of features of the form
\begin{equation}\label{LinComb}
\sum_{k=1}^{p} a_{k}x_{k}+a_{p+1} \leq 0 \text {,  where }  a_1, a_2, \dots , a_{p+1} \in \mathbb{R}.
\end{equation}
One of the major issues when inducing an oblique DT is the time complexity of the induction algorithm. In a data structure with $p$ feature variables and $n$ examples at a non-terminal node, the number of splits to be evaluated to find the best axis parallel split is $O(np)$. Therefore, the globally optimal split (with respect to an impurity function) at a non-terminal node can be found by exhaustively searching all possible splits along the feature axes. However, the number of splits to be evaluated to find the best oblique split at a node by exhaustive search is at most $O\left(2^p\times{n \choose p}\right)$ \citep{murthy1994system}. Hence, an  exhaustive search for the best oblique split is impractical. Furthermore, the best split at a node does not necessarily lead to the optimal tree. Spending more time searching for the best split at a node in general may not be beneficial \citep{heath1993induction}. Furthermore, \citep{hyafil1976constructing} point out the problem of finding an optimal binary DT is an NP-complete problem. This led us to search for efficient heuristics for constructing near optimal decision trees. In this work, we propose a simple, and effective heuristic method to induce oblique decision trees. \\
The remaining sections of this paper are organized as follows: Section~\ref{Lit_Rwv} highlights related work. Section~\ref{Prop_Mthd} introduces our proposed method. Comparisons with some commonly used DT algorithms are presented in Section~\ref{Exp_Res}. Section~\ref{Discus} concludes the paper with discussions.
\section{Related Work}\label{Lit_Rwv}
Most of the oblique DT induction algorithms construct DTs in a top-down fashion \citep{rokach2005top}. The induction algorithms differ in the way they search for the best split and can be categorised as follows. We define three categories: Induction algorithms that use optimisation techniques;  standard statistical techniques; and those that use heuristic arguments.\\
\subsection{Tree induction methods based on optimisation techniques}\label{OptDts}
The first major oblique DT algorithm was Classification and Regression Trees - Linear Combination, which is commonly known as CART-LC \citep{breiman1984classification}. CART-LC uses a deterministic hill climbing algorithm to search for the best oblique split at a non-terminal node. A backward feature elimination process is also carried out to delete irrelevant features from the split. CART-LC will not necessarily find the best split at each node because there is no built in mechanism to avoid getting stuck in the local maxima of $\Delta(I)$. The best split found may be only a local, rather than global, maximiser of $\Delta(I)$. \\
 
Simulated Annealing Decision Tree (SADT) was introduced by \citep{heath1993induction}. This DT uses the simulated annealing optimisation algorithm, which uses randomisation, to search for the best split. The use of randomisation potentially avoids getting stuck in local maxima of $\Delta(I)$ and will often produce better trees than those of CART-LC. The main disadvantage of the algorithm is the time taken to find the best split. In some cases it may require the evaluation of tens of thousands of hyperplanes before finding an optimal split \citep{murthy1994system}. \\
 The concepts of CART-LC and SADT are combined to produce a new oblique DT methodology called OC1 by \citep{murthy1994system}. Their method uses a deterministic hill climbing algorithm to perturb the coefficients of an initial hyperplane until a local maximum of $\Delta(I)$ is found. Then the hyperplane is perturbed randomly in an attempt to find a hyperplane that improves  $\Delta(I)$ further. These two steps are repeated several times. Each time the algorithm starts with a different initial hyperplane, with one being the best axis parallel split and the others chosen randomly. After many hyperplanes have been evaluated, the one that maximises the $\Delta(I)$ is taken as the splitting hyperplane. The time complexity at each non-terminal node for OC1 in the worst case scenario is shown to be $O\left(pn^2\log(n)\right)$ provided that Max Minority or Sum Minority impurity measures are used. However, the complexity increases for other impurity measures and for multi-class problems. One feature of both SADT and OC1 is that both algorithms can construct different decision trees on different runs using the same learning sample. Therefore, it is possible to run these algorithms multiple times and pick the best tree. However, this advantage is only realised on relatively small training example sets.
\subsection{Tree induction methods based on standard statistical techniques}
Various oblique DT induction algorithms have been developed using standard statistical techniques and can be found in \citep{gama1999linear},\cite{kolakowska2005fisher} ,\citep{li2003multivariate}   and \cite{lopez2013fisher}. The advantage of this approach is that the time required to induce DTs is generally lower than those based on optimisation algorithms. 
Quick Unbiased Efficient Statistical Tree (QUEST) \citep{loh1997split} uses Linear Discriminant Analysis (LDA) to find the best split at each node and hence there is no requirement for searching for the best split. QUEST's axis parallel tree begins by performing an ANOVA test at each non-terminal node to select the best feature. LDA is then applied on the selected feature to find the best splitting point. QUEST's oblique DT simply applies LDA on all features to find the best splitting hyperplane. Furthermore, QUEST is able to find oblique splits which are a linear combination of qualitative and quantitative  features. For multi-class problems, QUEST groups the classes into two super-classes using 2-means clustering algorithm and this increases the time complexity of the algorithm. \\

\subsection{Tree Induction Methods based on Heuristics}
DTs based on heuristic arguments have gained more popularity in recent past. In this approach, a logic is constructed by assuming  structure of class boundaries. If the assumption is true, DTs based on heuristic arguments produce accurate and smaller trees. DTs based on heuristic arguments can be found in  \citep{amasyah2008cline}   and \citep{manwani2012geometric}.

The CARTopt algorithm introduced by \citep{robertson2013cartopt}, uses a two class oblique tree to find a minimiser of a nonsmooth function $f(\bf{x})$ where $\bf{x}$ $\in \mathbb{R}^n$. Initially the examples in $\mathbb{R}^n$ are labelled as high and low depending on their value of $f(\bf{x})$.  An oblique DT is then used to form a partition on $\mathbb{R}^n$ which separates the low points from high points. Rather than forming the oblique DT directly, the authors reflected the training examples using a Householder matrix. Axis parallel splits are then searched in the reflected training data. These splits are oblique in original space.\\
CARTopt introduces a new heuristic to induce oblique decision trees. It uses the simplest form of splits, axis parallel splits,  to find oblique splits. Hence time complexity of searching oblique splits using CARTopt's approach is less than those based on optimisation algorithms. In this study we extend the CARTopt's idea in a number of ways to develop a complete oblique DT for statistical data classification. \\

\section{Methodology}\label{Prop_Mthd}
\begin{figure}[!ht]
\begin{center}
\includegraphics[width=.7\textwidth,height=.4\textwidth]{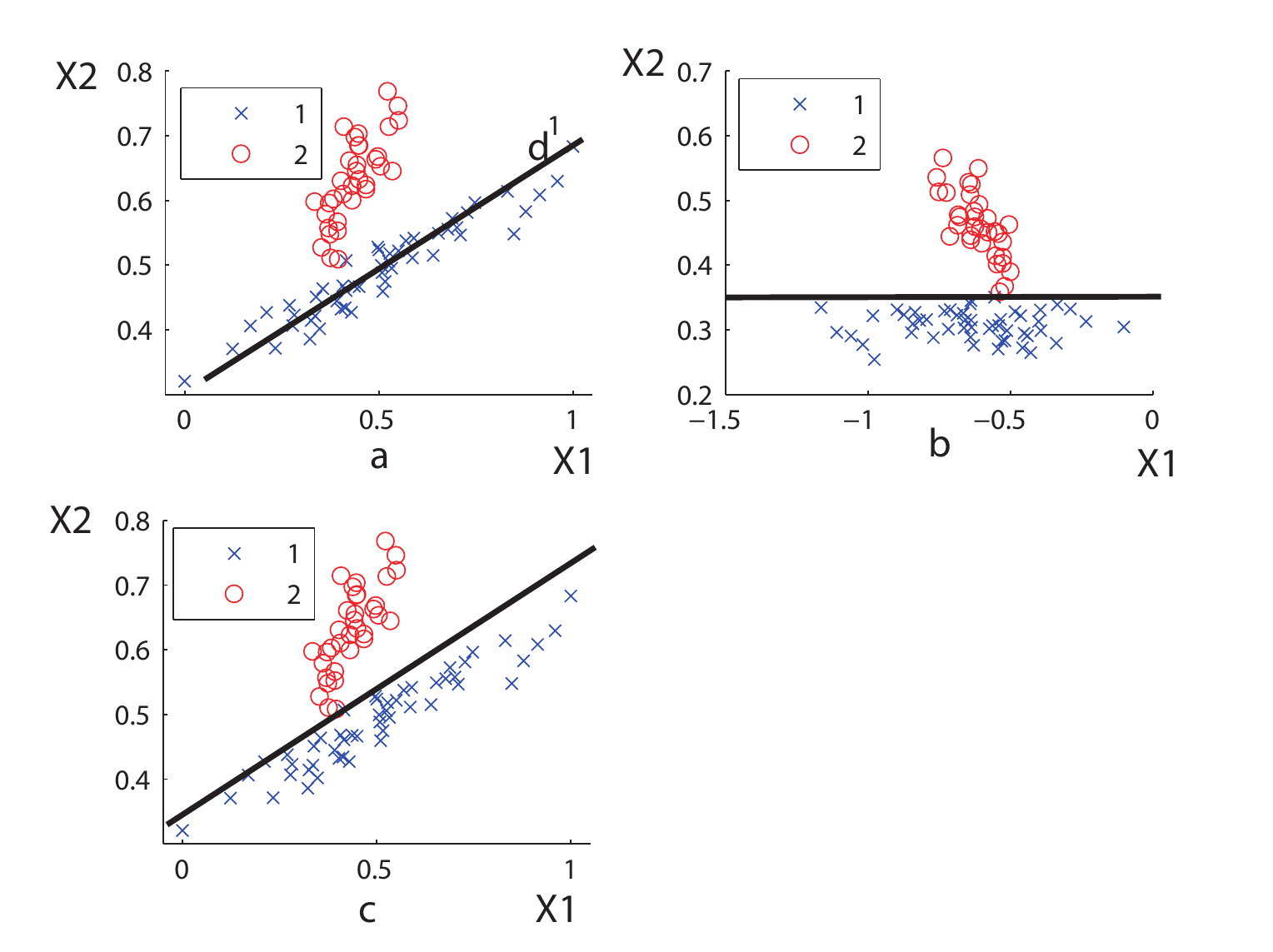} 
\caption{\small{Mechanism of the Householder Reflection. a). Scatter in the original space. $d^1$ is the dominant eigenvector of the class covariance matrix of class one. b). Scatter in the reflected space and the best axis parallel split found. c). Oblique split in the original space. }}
\label{fig:Fig1}
\end{center}
\end{figure}

We extend the oblique DT method used in the CARTopt optimisation algorithm of \citep{robertson2013cartopt} in a number of ways to develop a complete oblique DT called HHCART. First, CARTopt is designed to classify two classes whereas HHCART can handle multi-class classification problems. Second, CARTopt reflects the training examples at the root node only whereas  HHCART performs reflections at each non-terminal node during tree construction. Finally, \mbox{CARTopt} is only defined for quantitative features whereas HHCART is capable of finding oblique splits which can be linear combinations of both quantitative and qualitative features.\\ 
First, we explain the basic concept of our algorithm for a two class classification problem. The algorithm easily generalises to the multi-class problem. In our approach we find each separating hyperplane by considering the orientation of each class. We propose the dominant eigenvector of the covariance matrix of a class to represent the orientation of that class. If this orientation is parallel to one of the feature axes, the best separating hyperplane may be found by performing axis parallel splits. Otherwise, we reflect the set of examples to a new coordinate system such that the orientation of one of the classes becomes parallel to one of the axes in the reflected feature space. Axis parallel splits can then be searched in the reflected feature space to find the best split. This split will be oblique in the original feature space \citep{robertson2013cartopt}.\\
Consider the two dimensional, two class classification problem shown in \mbox{Figure~\ref{fig:Fig1} (a)}.\\
First we define the estimated covariance matrix of a set of examples. Let $\bf{x}_\mathrm{1}, \bf{x}_\mathrm{2}, \dots ,\bf{x}_\mathrm{n}$ be $p$ dimensional feature vectors where \mbox{$\bf{x}_\mathrm{i}$ $= (\it{x}_{i1}, \it{x}_{i2}, \dots, \it{x}_{ip})^T$}. Then the estimated covariance matrix is given by\\
$S=\frac{1}{(n-1)}\sum_{i=1}^{n}{(\bf{x}_\mathrm{i} - \bf{\bar{x}})(\bf{x}_\mathrm{i} - \bar{\bf{x}})}$$^{T}$ where \mbox{$\bar{\bf{x}} = (\bf{\bar{\it{x}}_\mathrm{1}}, \bf{\bar{\it{x}}_\mathrm{2}}, \dots, \bf{\bar{\it{x}}_\mathrm{p}})$$^T$} is the mean vector. \\
We reflect the examples using a Householder matrix which can be defined as follows. Let $\bf{d}^\mathrm{1}$ be the dominant eigenvector of the estimated covariance matrix of class 1 examples.  Then there exists an orthogonal symmetric matrix $\textbf{H}_{p\times p}$ (where $p$ is number of features) such that
\begin{equation}\label{HHolder}
\begin{split}
    H=I-2{\bf{u}\bf{u}}^{T} \text{   where   } \bf{u}=\frac{\bf{e}_\mathrm{1}-\bf{d}^\mathrm{1}}{\|\bf{e}_\mathrm{1}-\bf{d}^\mathrm{1}\|_\mathrm{2}} \\ \text{ and } {\bf{e}_\mathrm{1}}_{p \times 1}=(1,0, \dots, 0)^\mathrm{T}.
\end{split}
\end{equation}
Let $\mathfrak{D}_{n\times p}$ be the training example set. The reflected example set $\hat{\mathfrak{D}}_{n\times p}$ is obtained using $\hat{\mathfrak{D}} = \mathfrak{D}H$. Since $H_{p\times p}$ is symmetric and orthogonal, a point in the transformed space can be mapped back to original space at a minimal cost ($HH=I$). The mechanism of the Householder reflection is that it reflects vector $d^1$ on to $e_1$ by a reflection through the plane perpendicular to vector $e_1-d^1$. The reflected example set is shown in \mbox{Figure~\ref{fig:Fig1} (b)}. \\
Each column of $H$ represents the direction of a coordinate axis in the reflected space. Axis parallel splits are searched along these axes. These splits are oblique in the original space. The best axis parallel split found in the reflected space, which is oblique in the original space, is shown in Figure~\ref{fig:Fig1} (c).\\
The axis parallel search space can be enhanced by using all possible eigenvectors for reflections. For a $p$-dimensional classification problem with $C$ classes there are $Cp$ eigenvectors to be considered for the Householder reflection. However, this increases the time complexity of tree induction, but have an opportunity to produce better trees. \\
Here we explain the complete algorithm of HHCART. We propose two versions of HHCART: HHCART(A) is based on all possible eigenvectors of all classes and HHCART(D) is based on only the dominant eigenvector of each class. For any given non-terminal node $t$, let$\mathfrak{D}_t$ and $C_t$ be the set of examples and classes available at that node respectively. At node $t$, HHCART(A) finds all eigenvectors of the estimated covariance matrix for each class whereas HHCART(D) finds only the dominant eigenvector of each class. A Householder matrix is constructed for each eigenvector. Then $\mathfrak{D}_t$is reflected using each Householder matrix, and axis parallel splits are performed along each coordinate axis in the reflected space. The best axis parallel split is chosen as the separating hyperplane at node $t$.  However, if the eigenvector is already parallel to any of the feature axes, no reflection is done and hence axis parallel splits are searched in the original space. The hyperplane found divides node $t$ into two child nodes. The algorithm is recursively run on all child nodes until each child node satisfies either: 
\begin{enumerate}
\item[a.]  the misclassification rate at the child node is either 0 or not greater than a user specified threshold (MisRate); or
\item[b.] the number of examples in the node is less than or equal a user specified threshold \mbox{(MinParent)}.

\end{enumerate}
An overview of HHCART(A) algorithm at node $t$ is given in Algorithm~\ref{alg:Algo1}. The time complexity at a node for HHCART(A) in the worst case is $O(Cn^2p^3)$ (See appendix A for the derivation). However, if HHCART(D) is used the time complexity reduces to $O(Cn^2p^2)$.
\begin{algorithm}[!t]
\caption{Overview of HHCART(A) algorithm at a single node}\label{alg:Algo1}
\SetAlgoLined
\KwData{Input: Examples at node $t$, called $\mathfrak{D}_t$,  Minparent, MisRate, and $\tau > 0$. \\}
 initialization\;
Define $N_t$ = Number of examples in $\mathfrak{D}_t$\;
Define $mp_t$ = misclassification rate at node $t$\;
Define $C_t$ = number of classes at node $t$\;
Define $p$ = number of features\;
$\Delta(I_{max}) = 0$\;
$h_t = empty$\;
    \If {($N_t > \text{Minparent}$) and ($\text{MisRate} <  mp_t$)}
    {
    \For {i=1:$C_t$}
     {
     Extract the the examples that belong to the $i^{th}$ class in $\mathfrak{D}_t$, called $D_i$\;
     Compute the normalized eigenvectors and eigenvalues  of estimated covariance matrix for $D_i$\;\
					 \centerline{(($d^{1i}, \lambda^{1i}$), \dots ($d^{pi},\lambda^{pi}$))}\ 
      \For {j=1:p}
      {
	 \If {$\lambda^{ji} \ne 0$}
		{
		\eIf {$\|e_1 - d^{ji} \| \leq \tau$ or $\|e_2 - d^{ji} \| \leq \tau$ or $\dots$ or $\|e_p - d^{ji} \| \leq \tau$ } 
		{
			$H^{ji}_t = I$, the Identity matrix\;
		}
		{
     		Construct the Householder matrix $H^{ji}_t$ using $d^{ji}$ \;	
		}
       Reflect $\mathfrak{D}_t$ : $\hat{\mathfrak{D}}_{t} = \mathfrak{D}_t*H^{ji}_t $\;
       Find the best axis parallel hyperplane split, called $h^{ji}_{t}$\;
      \If {$\text{impurity reduction of } h^{ji}_{t} > \Delta(I_{max})$}
        {
        Replace $h_t$ with $h^{ji}_{t}$, the best hyperplane found so far\;
		Replace $\Delta(I_{max})$ with the impurity reduction of $h^{ji}_t$
         }
        }
        }
       }
}
\end{algorithm}
\subsection{Small Samples}
As the tree grows, the number of examples at each node usually becomes small. This raises two questions to be answered. (a) Is it worthwhile searching for an oblique split or is an axis parallel split sufficient? (b) How are the eigenvectors calculated for small sample sizes? The first problem is common for any oblique DT. In the OC1 algorithm the authors suggest using oblique splits if the number of examples at a node is greater than twice the number of feature variables. The second question has two parts:
    \begin{center}
    \begin{enumerate}
        \item[1] Non-availability of some eigenvalues (the ones equal to zero) due to a singular covariance matrix. 
        \item[2] Performing an eigen analysis for classes having only one example or several examples with the same feature vector.
    \end{enumerate}
    \end{center}
Part (1)  can be solved without modifying the HHCART algorithm because the reflection is done using the eigenvectors whose eigenvalues are not zero. For part (2), we simply omit classes that have a single example or several examples with the same feature vector. However, if all the classes suffer from this problem, then axis parallel splits are performed.
\subsection{Qualitative Variables}
Data classification problems often contain a mixture of quantitative and qualitative feature variables. Since the class discriminatory information may be contained in both types of feature variables, an effective classifier should be able to handle both types of features in the classification process. For a qualitative feature variable $X$, the form of the split is given by $X \in A$ where $A$ is a non-empty subset of values taken by $X$. If a qualitative feature has $M$ non-empty levels, $2^{M-1}-1$ splits are possible. Axis parallel algorithms which consider qualitative splits can be found in \citep{quinlan1986induction}.\\
Incorporating qualitative features in oblique splits has not been explored much. The QUEST algorithm \citep{loh1997split} is capable of finding oblique splits with both qualitative and quantitative features. QUEST transforms each unordered qualitative feature variable into a new ordered quantitative feature variable. Each level of an unordered qualitative feature is mapped to a ordered value called a CRIMCOORD. The exact CRIMCOORD algorithm can be found in \citep{loh1997split}.
We implement the same CRIMCOORD algorithm in HHCART to induce oblique splits which contain both qualitative and quantitative features. At each node, a new quantitative feature is constructed for each qualitative feature by mapping its levels to CRIMCOORDS. Then these new quantitative features are amalgamated with the existing quantitative features in the example set. The HHCART algorithm can then be applied to find the best oblique split. At each node the CRIMCOORD corresponding to each level of each qualitative feature is stored. When predicting, the level of each qualitative feature of an unclassified observation is replaced by the corresponding CRIMCOORD attached to each node along its path. 

\section{Experiments}\label{Exp_Res}
Two sets of experiments are carried out to compare the performance of HHCART  with other DT methods. The first experiment considers quantitative example sets and the second experiment considers example sets with both qualitative and quantitative features. Both HHCART(A) and HHCART(D) methods are considered in the experiments.
\subsection{Comparison on datasets having quantitative features only }\label{Exp1}
In this section, we compare the HHCART methods with OC1, OC1-LC (OC1 version of Breiman's linear combination
methods) and OC1-AP (OC1 version of axis parallel splits). All of these methods are available in the OC1 system which is freely available at \citep{murthyoc1}. However, the backward feature elimination process of Breiman's CART-LC method is not included in OC1-LC and hence is somewhat different from the original method.
\subsubsection{Experimental Setup}
Experiments were performed on real data sets that were downloaded from \citep{Bache+Lichman:2013} and are given in Table~\ref{tab:table1}. In our algorithm we set MinParent=2, MisRate=0 and $\tau =0.05$. For OC1, OC1-LC and OC1-AP MinParent was set to 2. All the algorithms used the Twoing rule as the measure of impurity \citep{breiman1984classification} and  Cost complexity pruning \citep{breiman1984classification} with zero standard error. For OC1, the number of restarts and number of jumps were set to 20 and 5 (default values) respectively. Five-fold cross validations  were used to estimate the classification accuracy. For each fold,  $10\%$ of the training set was used exclusively for pruning. We then used ten, five-fold cross validations to estimate the accuracy and the size of the tree. Therefore, to estimate accuracy and tree size (number of terminal nodes) the average over ten runs was used.  Results are reported in Table~\ref{tab:table2} along with respective standard deviations.
The Shuttle data set comes with its own training set containing  43500 examples and a test set with 14500 examples. Therefore instead of performing a cross validation experiment, we induced $10$ trees, each using $90\%$ of training examples for induction and remaining $10\%$ for pruning. The accuracy of all the trees was estimated using the Shuttle data test set. Since approximately $80\%$ of the examples belong to class 1, the aim is to achieve an accuracy between $99-99.9\%$ \citep{Bache+Lichman:2013}.\\
\begin{table}[!htbp]
\caption{Real Data sets with quantitative features, downloaded from UCI Repository}
\label{tab:table1}
\begin{center}
\begin{tabular}{ccccc}
\hline\noalign{\smallskip}
  Data set  & No. of &  No. of & No. of  \\
            &   feature & classes & examples\\  
  \noalign{\smallskip}\hline\noalign{\smallskip}
  Heart (HRT) & 13 & 2& 270            \\
  Pima Indian (PIND)   & 8    & 2& 768            \\
  Breast Cancer (BC)  & 9    & 2& 638            \\
  Boston Housing (BH)  & 13    & 2& 506            \\
  Wine(WINE)   & 13    & 3& 178            \\
  BUPA  & 6    & 2& 345            \\
  Balance Scale (BS)  & 4    & 3& 625            \\
  Glass (GLS)  & 9    & 7& 214            \\
  Shuttle (SHUT)  & 9    & 7& 58000            \\
  Letter (LET)   & 10    & 26& 20000            \\
  Survival (SUR)  & 3    & 2& 306           \\
 \noalign{\smallskip}\hline
  \end{tabular}
 \end{center}
\end{table}

\begin{table}[!h]
\caption{Results of HHCART and other DT methods}
  \label{tab:table2}
  \centering
    \scalebox{0.8}{
    \begin{tabular}{llccllcc}
    \hline \noalign {\smallskip}
    \textbf{Dataset} & \textbf{DT} & \textbf{Avg. Acc.} & \textbf{Avg. Size} & \textbf{Dataset} & \textbf{DT} & \textbf{Avg. Acc.} & \textbf{Avg. Size} \\
     \noalign{\smallskip}\hline\noalign{\smallskip}
    BS & HHCART(A) & $\textbf{93.7}\pm1.3$  & $7.9\pm1.7$ & PIND & HHCART(A) & $72.2\pm2.0$  & $9.1\pm5.1$ \\
       & HHCART(D) & $88.3\pm1.7$  & $12.2\pm3.5$ &   & HHCART(D) & $72.9\pm1.3$  & $10.8\pm4.4$ \\
          & OC1   & $91.9\pm0.9$  & $8.7\pm3.4$ &       & OC1   & $73.4\pm1.0$  & $9.2\pm5.4$ \\
          & OC1-AP & $78.2\pm1.3$  & $37.5\pm16.8$ &       & OC1-AP & $\textbf{73.6}\pm1.4$  & $15.9\pm8.7$ \\
          & OC1-LC & $84.3\pm1.5$  & $12.6\pm6.5$ &       & OC1-LC & $72.8\pm1.8$  & $11.4\pm9.6$ \\ \hline
    BH & HHCART(A) & $\textbf{83.3}\pm0.9$      & $6.5\pm2.1$ & SHUT & HHCART(A) & $99.94\pm0.02$      & $25.4\pm5.9$ \\
	   & HHCART(D) & $83.0\pm0.7$      & $9.9\pm2.6$ &              & HHCART(D) & $99.94\pm0.05$      & $26.1\pm4.9$ \\
          & OC1   & $82.2\pm1.2$       & $9.3\pm3.4$ &       & OC1   &   $99.95\pm0.03$    & $32.6\pm7.71$ \\
          & OC1-AP &$82.0\pm0.7$       & $13.0\pm5.3$ &       & OC1-AP & $\textbf{99.97}\pm0.02$      & $26.5\pm5.6$ \\ 
          & OC1-LC &  $81.5\pm1.3$     & $10.6\pm6.0$ &       & OC1-LC &  $88.4\pm7.07$     & $44.7\pm42.4$ \\ \hline
    BC & HHCART(A) & $\textbf{97.0}\pm0.3$      & $2.4\pm0.6$ & WINE  & HHCART(A) & $\textbf{91.3}\pm1.6$      & $3.4\pm0.3$ \\
       & HHCART(D) & $\textbf{97.0}\pm0.3$      & $2.6\pm1.1$ &       & HHCART(D) & $88.7\pm3.1$      & $4.5\pm0.6$ \\
          & OC1   & $95.4\pm0.5$      & $3.3\pm1.4$ &       & OC1   &   $89.2\pm2.1$    & $3.5\pm0.3$ \\
          & OC1-AP & $94.0\pm0.8$      & $8.3\pm3.3$ &       & OC1-AP &  $89.2\pm4.6$     & $4.6\pm0.6$ \\ 
          & OC1-LC & $95.5\pm0.6$      & $3.4\pm1.6$ &       & OC1-LC &  $89.4\pm2.7$     & $3.8\pm0.6$ \\ \hline
    BUPA  & HHCART(A) &  $64.1\pm2.6$     & $6.5\pm1.5$ & LET & HHCART(A) &   $82.1\pm0.3$    & $759.2\pm88.1$ \\
          & HHCART(D) &  $62.4\pm2.5$     & $8.6\pm3.1$ &     & HHCART(D) &   $83.1\pm0.3$    & $1135.9\pm122$ \\
          & OC1   & $\textbf{66.9}\pm2.2$      & $8.9\pm6.1$ &       & OC1   &  $83.6\pm0.4$     & $1197.2\pm88.9$ \\
          & OC1-AP & $64.7\pm2.5$      & $13.2\pm10.5$ &       & OC1-AP & $\textbf{86.3}\pm0.3$      & $1611.7\pm60.0$ \\ 
          & OC1-LC & $64.4\pm2.4$      & $8.9\pm3.6$ &       & OC1-LC &  $84.5\pm0.2$     & $1332.6\pm146.3$ \\ \hline
    GLS & HHCART(A) & $60.3\pm3.0$      & $8.5\pm3.0$ & SUR & HHCART(A) &  $\textbf{73.5}\pm1.5$     & $5.3\pm2.7$ \\
        & HHCART(D) & $61.9\pm3.0$      & $10.1\pm2.3$ &  &  HHCART(D) &  $72.8\pm1.0$     & $5.0\pm2.4$ \\
          & OC1   & $61.1\pm3.5$      & $10.8\pm4.3$ &       & OC1   &    $71.0\pm2.1$   & $6.4\pm3.5$ \\
          & OC1-AP & $64.6\pm3.9$      & $14.6\pm8.7$ &       & OC1-AP &   $71.9\pm1.5$    & $10.7\pm6.5$ \\
          & OC1-LC & $\textbf{67.4}\pm2.0$      & $12.0\pm3.6$ &       & OC1-LC &  $70.2\pm2.4$     & $8.1\pm4.4$ \\ \hline
    HRT & HHCART(A) &  $74.1\pm2.9$     & $4.5\pm1.7$ &       &       &       &  \\
        & HHCART(D) &  $75.8\pm2.8$     & $7.8\pm2.6$ &       &       &       &  \\
          & OC1   &  $\textbf{77.1}\pm2.5$     & $3.6\pm1.0$ &       &       &       &  \\
          & OC1-AP & $76.3\pm2.3$      & $6.7\pm2.4$ &       &       &       &  \\
          & OC1-LC & $76.3\pm2.5$      & $4.0\pm1.1$ &       &       &       &  \\
    \noalign{\smallskip}\hline
    \end{tabular}
  }\end{table}%

Table~\ref{tab:table2} shows the results for our first experiment.  The average accuracies and the average tree sizes of ten, five-fold cross classifications are listed in the table. It is clear that oblique splits reduce the average tree size for all the data sets while increasing the accuracy for most of data sets. The average accuracy of HHCART(A) is significantly (more than 2 standard deviations) higher than  all the other methods tested for the BC dataset. Also the average accuracy of HHCART(A) is higher than the other methods for BS, BH, WINE and SUR datasets. \\
The average tree sizes of HHCART(A) are consistently smaller than the average tree sizes of other methods except for the HRT dataset. Therefore the performance of HHCART(A) with respect to accuracy and tree size is better than the other methods for BS, BH, BC, WINE and SUR datasets. \\
Eight of the eleven datasets have at least 8 features. For six of these relatively high dimensional data sets, the performance of HHCART(A) is comparable with OC1 and OC1-LC. Therefore we can conclude that the proposed method works well in relatively high dimensional feature spaces.\\
 For all the datasets except BS and WINE,  HHCART(D) performs as well as HHCART(A) in terms of the average accuracy. Also the tree sizes of HHCART(D) are comparable with those produce by HHCART(A) except for the BH, BUPA and HRT datasets. The performance of HHCART(D) is as similar as OC1 with respect to both the accuracy and tree size for all the datasets except the BS, HRT and BUPA datasets.    
The time complexity of HHCART(A) is higher than that of HHCART(D) by a factor of $O(p)$. Results show that HHCART(D) produces DTs with similar accuracies and sizes as HHCART(A) and OC1 for most of the datasets. Hence HHCART(D) would be a more efficient method to use for higher dimensional problems.
\subsection{Comparison on Datasets having qualitative and quantitative features}\label{Exp2}
Experiments were performed to study the performance of the HHCART methods when the training examples contain both qualitative and quantitative features. Since  OC1, OC1-AP, and OC1-LC are not designed to handle oblique splits containing both qualitative and qualitative features, QUEST \citep{loh1997split} was used for comparison purposes.
\subsubsection{Experimental setup}
Experiments were performed on the datasets available in \citep{Bache+Lichman:2013}, which are given in Table~\ref{tab:table3}. Ten, five-fold cross validations were used and the average accuracies and tree sizes (over ten cross validations) are reported in  Table~\ref{tab:table4}. The Income dataset comes with its own training and testing set of 30162 and 15060 examples respectively. We induced $10$ trees, each using $90\%$ of the training examples and the remaining $10\%$ were used for pruning. The accuracy of all the trees were estimated using the same test set. 
\begin{table}[htbp]
\caption{Real Data sets with qualitative and quantitative features, downloaded from UCI Repository}
\label{tab:table3}
\begin{center}
\begin{tabular}{ccccc}
 \hline\noalign{\smallskip}
  Data Set & No. of features  & No. of & No. of \\
           &   (No. of Qualitative) &  Classes &  Examples\\   
\noalign{\smallskip}\hline\noalign{\smallskip}
  Income & 14(8)   & 2    & 45222          \\
  Bank  & 16(9)    & 2 & 45211            \\
  StatLog   & 14(8)    & 2& 690            \\
\noalign{\smallskip}\hline
  \end{tabular}
\end{center}
\end{table}
QUEST uses the following parameter setting: estimated priors, unit misclassification cost, zero standard error for pruning, linear splits, linear discriminant analysis for the split point, minimum node size for splitting =2. The HHCART methods were implemented as above.   
\begin{table}[htbp]
\caption{{Results of HHCART and QUEST }}
\label{tab:table4}
  \centering
    \begin{tabular}{llcc}
  \hline\noalign{\smallskip}
    \textbf{Dataset} & \textbf{Decision Tree} & \textbf{Avg. Acc.} & \textbf{Avg. Size}\\
  \noalign{\smallskip}\hline\noalign{\smallskip}
    Income & HHCART(A) & $85.1\pm0.2$  & $32.7\pm12.9$ \\
		   & HHCART(D) & $\textbf{85.5}\pm0.2$  & $59.5\pm19.7$ \\
          & QUEST   & $83.9\pm0.2$  & $68.0\pm23.1$ \\
   \hline
    Bank & HHCART(A) & $90.2\pm0.12$      & $22.58\pm11.94$ \\
	     & HHCART(D) & $\textbf{90.4}\pm0.07$      & $44.4\pm14.19$ \\
          & QUEST   & $90.1\pm0.1$       & $27.0\pm15.2$ \\
   \hline
    StatLog & HHCART(A) & $85.1\pm0.9$      & $5.6\pm1.9$ \\
	        & HHCART(D) & $\textbf{85.8}\pm0.7$      & $6.5\pm3.0$ \\
          & QUEST   & $85.65\pm0.92$      & $6.08\pm3.6$ \\
   \noalign{\smallskip}\hline
    \end{tabular}
\end{table}%
For the Income dataset, HHCART(A)'s performance is significantly (more than 2 standard deviations) better than QUEST both in terms of the average accuracy and average tree size. For the other two datasets, HHCART(A) produces comparable accuracies with smaller trees. These results also suggest that the HHCART algorithms perform well in relatively high dimensions. Though HHCART(D) produces larger trees compared with HHCART(A), its classification accuracy is comparable with HHCART(A).  

\section{Conclusions}\label{Discus}
In this work we have presented a new way of inducing oblique DTs called HHCART. It uses the eigenvectors of the estimated covariance matrices of respective classes to define a Householder matrix which is used to reflect the examples so that reflected axis parallel splits can be found. Two versions of HHCART have been presented: HHCART(A) uses all possible eigenvectors of the estimated covariance matrices of respective classes whereas HHCART(D) uses only the dominant eigenvector of each class.  Based on the empirical results obtained, it is clear that both HHCART methods perform well in terms of accuracy and tree size. Furthermore, HHCART is capable of classifying datasets with both qualitative and quantitative features.\\

 \appendix

\section{Time Complexity of HHCART}
Here we derive the maximal time complexity at a node of HHCART(A) and \mbox{HHCART(D)}. Assume there are $n$ examples with $p$ quantitative features and $C$ classes at the node.
\begin{enumerate}
   \item[1]  \textbf{HHCART(A)} and  \textbf{HHCART(D)}  - Complexity for constructing estimated covariance matrix for one class of examples is $O(np^2)$. For $C$ classes the complexity is $O(Cnp^2)$.
    \item[2]  \textbf{HHCART(A)} - Complexity of the complete eigen analysis for one class of examples is $O(p^3)$. For $C$ classes the complexity is $O(Cp^3)$.\\
    		  \textbf{HHCART(D)} - Complexity for finding the dominant eigenvector for one class of examples is $O(p^2)$. For $C$ classes the complexity is $O(Cp^2)$.
    \item[3]  \textbf{HHCART(A)} - Complexity for the reflection of $n$ examples using one Householder matrix is $O(np^2)$. Since there are $Cp$ Householder matrices the Complexity is $O(Cnp^3)$.\\
              \textbf{HHCART(D)} - Complexity for the reflection of $n$ examples using one Householder matrix is $O(np^2)$. 
              					   For $C$ Householder matrices the complexity is $O(Cnp^2)$. 		
     \item[4] \textbf{HHCART(A)} - Complexity of finding the best axis parallel splits for one reflected space is $O(n^2p)$. Since there are $Cp$ reflected spaces the Complexity is $O(Cn^2p^2)$. \\
         \textbf{HHCART(D)} -  Complexity of finding the best axis parallel splits for one reflected space is $O(n^2p)$. For $C$ classes the complexity is $O(Cn^2p)$ 
      \item[5] \textbf{HHCART(A)} - The maximal time complexity at a node is  $O(Cnp^2)$+$O(Cp^3) + O(Cnp^3) + O(Cn^2p^2) = O(Cn^2p^3)$.\\
    		   \textbf{HHCART(D)} - The maximal time complexity at a node is $O(Cnp^2)$+ $O(Cp^2) + O(Cnp^2) + O(Cn^2p) = O(Cn^2p^2)$.
\end{enumerate}




\bibliographystyle{elsarticle-harv}
\bibliography{RefList}

\begin{thebibliography}{18}
\expandafter\ifx\csname natexlab\endcsname\relax\def\natexlab#1{#1}\fi
\expandafter\ifx\csname url\endcsname\relax
  \def\url#1{\texttt{#1}}\fi
\expandafter\ifx\csname urlprefix\endcsname\relax\def\urlprefix{URL }\fi

\bibitem[{Amasyah and Ersoy(2008)}]{amasyah2008cline}
Amasyah, M., Ersoy, O., 2008. Cline: A new decision-tree family. Neural
  Networks, IEEE Transactions on 19~(2), 356--363.

\bibitem[{Bache and Lichman(2013)}]{Bache+Lichman:2013}
Bache, K., Lichman, M., 2013. {UCI} machine learning repository.
\newline\urlprefix\url{http://archive.ics.uci.edu/ml}

\bibitem[{Breiman et~al.(1984)Breiman, Friedman, Stone, and
  Olshen}]{breiman1984classification}
Breiman, L., Friedman, J., Stone, C.~J., Olshen, R.~A., 1984. Classification
  and regression trees. CRC press.

\bibitem[{Gama and Brazdil(1999)}]{gama1999linear}
Gama, J., Brazdil, P., 1999. Linear tree. Intelligent Data Analysis 3~(1),
  1--22.

\bibitem[{Heath et~al.(1993)Heath, Kasif, and Salzberg}]{heath1993induction}
Heath, D., Kasif, S., Salzberg, S., 1993. Induction of oblique decision trees.
  In: IJCAI. Citeseer, pp. 1002--1007.

\bibitem[{Hyafil and Rivest(1976)}]{hyafil1976constructing}
Hyafil, L., Rivest, R.~L., 1976. Constructing optimal binary decision trees is
  np-complete. Information Processing Letters 5~(1), 15--17.

\bibitem[{Ittner and Schlosser(1996)}]{ittner1996non}
Ittner, A., Schlosser, M., 1996. Non-linear decision trees-ndt. In: ICML.
  Citeseer, pp. 252--257.

\bibitem[{Kolakowska and Malina(2005)}]{kolakowska2005fisher}
Kolakowska, A., Malina, W., 2005. Fisher sequential classifiers. IEEE
  Transactions on Systems, Man, and Cybernetics, Part B: Cybernetics 35~(5),
  988--998.

\bibitem[{Li et~al.(2003)Li, Sweigart, Teng, Donohue, Thombs, and
  Wang}]{li2003multivariate}
Li, X.-B., Sweigart, J.~R., Teng, J.~T., Donohue, J.~M., Thombs, L.~A., Wang,
  S.~M., 2003. Multivariate decision trees using linear discriminants and tabu
  search. IEEE Transactions on Systems, Man and Cybernetics, Part A: Systems
  and Humans 33~(2), 194--205.

\bibitem[{Li et~al.(2005)Li, Dong, and Kothari}]{li2005classifiability}
Li, Y., Dong, M., Kothari, R., 2005. Classifiability-based omnivariate decision
  trees. IEEE Transactions on Neural Networks 16~(6), 1547--1560.

\bibitem[{Loh and Shih(1997)}]{loh1997split}
Loh, W.-Y., Shih, Y.-S., 1997. Split selection methods for classification
  trees. Statistica sinica 7~(4), 815--840.

\bibitem[{L{\'o}pez-Chau et~al.(2013)L{\'o}pez-Chau, Cervantes,
  L{\'o}pez-Garc{\'\i}a, and Lamont}]{lopez2013fisher}
L{\'o}pez-Chau, A., Cervantes, J., L{\'o}pez-Garc{\'\i}a, L., Lamont, F.~G.,
  2013. Fisher’s decision tree. Expert Systems with Applications 40~(16),
  6283--6291.

\bibitem[{Manwani and Sastry(2012)}]{manwani2012geometric}
Manwani, N., Sastry, P., 2012. Geometric decision tree. IEEE Transactions on
  Systems, Man, and Cybernetics, Part B: Cybernetics 42~(1), 181--192.

\bibitem[{Murthy et~al.(1993)Murthy, Kasif, and Salzberg}]{murthyoc1}
Murthy, S.~K., Kasif, S., Salzberg, S., 1993. The oc1 decision tree software
  system.
\newline\urlprefix\url{http://www.cs.jhu.edu/salzberg/announce-oc1.html}

\bibitem[{Murthy et~al.(1994)Murthy, Kasif, and Salzberg}]{murthy1994system}
Murthy, S.~K., Kasif, S., Salzberg, S., 1994. A system for induction of oblique
  decision trees. J Artif. Intell Res. 2~(1), 1--32.

\bibitem[{Quinlan(1986)}]{quinlan1986induction}
Quinlan, J.~R., 1986. Induction of decision trees. Machine learning 1~(1),
  81--106.

\bibitem[{Robertson et~al.(2013)Robertson, Price, and
  Reale}]{robertson2013cartopt}
Robertson, B., Price, C., Reale, M., 2013. Cartopt: a random search method for
  nonsmooth unconstrained optimization. Computational Optimization and
  Applications 56~(2), 291--315.

\bibitem[{Rokach and Maimon(2005)}]{rokach2005top}
Rokach, L., Maimon, O., 2005. Top-down induction of decision trees
  classifiers-a survey. IEEE Transactions on Systems, Man, and Cybernetics,
  Part C: Applications and Reviews 35~(4), 476--487.

\end{thebibliography}







\end{document}